\documentclass[11pt]{article}
\usepackage{authblk}
\usepackage{coling2016}
\usepackage{times}
\usepackage{url}
\usepackage{latexsym}
\usepackage{graphicx}
\usepackage{caption}
\usepackage{subcaption}
\usepackage{wrapfig}

\usepackage{floatrow}
\newfloatcommand{capbtabbox}{table}[][\FBwidth]

\usepackage{color,soul}

\newcommand{\hawalay}{\emph{\d{h}aw\={a}lay}}

\title{Shamela: A Large-Scale Historical Arabic Corpus}

\author[1]{Yonatan Belinkov}
\author[2]{Alexander Magidow}
\author[3]{Maxim Romanov}
\author[46]{Avi Shmidman}
\author[56]{Moshe Koppel}
\affil[1]{MIT Computer Science and Artificial Intelligence Laboratory, Cambridge, MA, USA}
\affil[2]{Department of Languages and Literatures, University of Rhode Island, USA}
\affil[3]{The Humboldt Chair of Digital Humanities, Leipzig University, Leipzig, Germany}
\affil[4]{Department of Hebrew Literature, Bar-Ilan University, Israel}
\affil[5]{Department of Computer Science, Bar-Ilan University, Israel}
\affil[6]{Dicta: The Israel Center for Text Analysis}
\affil[ ]{}
\affil[ ]{{\tt belinkov@mit.edu, amagidow@uri.edu, maxim.romanov@uni-leipzig.de}}
\affil[ ]{{\tt shmidman@gmail.com, moishk@gmail.com}}

\date{}

\begin{document}
\maketitle
\vspace{40pt}
\begin{abstract}
Arabic is a widely-spoken language with a rich and long history spanning more than fourteen centuries. Yet existing Arabic corpora largely focus on the modern period or lack sufficient diachronic information. We develop a large-scale, historical corpus of Arabic of about 1 billion words from diverse periods of time. We clean this corpus, process it with a morphological analyzer, and enhance it by detecting parallel passages and automatically dating undated texts. We demonstrate its utility with selected case-studies in which we show its application to the digital humanities.  
\end{abstract}

\blfootnote{
   \hspace{-0.65cm}
   This work is licenced under a Creative Commons
   Attribution 4.0 International License.
   License details:\\
   \url{http://creativecommons.org/licenses/by/4.0/}
}

\section{Introduction}
\label{sec:intro}


Arabic has been used as a written language for more than fourteen centuries. While Arabic  attracts significant interest from the natural language processing (NLP) community, leading to large corpora and other valuable resources, mostly in Modern Standard Arabic (MSA),  we still lack a \textit{large-scale}, \textit{historical} corpus of Arabic that covers this entire time period. This lacuna affects three research communities: it hinders linguists from making corpus-driven historical analyses of the Arabic language; it prevents digital humanities (DH) scholars from investigating the history and culture of the Arabic-speaking people; and it makes it difficult for NLP researchers to develop applications for texts from specific historical periods. 

We aim to close this gap by developing a large-scale historical Arabic corpus. Our corpus is drawn from the Al-Maktaba Al-Shamela website,\footnote{\url{http://shamela.ws}} a website of Arabic texts from the early stages of the language (7th century) to the modern era. We have cleaned these texts and organized useful metadata information about them in a semi-automatic process. 
We have also lemmatized the entire corpus to facilitate semantic analysis. This step is especially important given the rich morphology of the Arabic language. The result is a corpus of over 6,000 texts, totaling around 1 billion words, of which 800 million words are from dated texts. We describe these procedures in some detail to facilitate future similar work. We present several case-studies showing how this corpus can be used for digital humanities research. The corpus itself will be made available to the research community.\footnote{An initial version is available in the RAWrabica collection: \url{https://github.com/OpenArabic/RAWrabica}.}


Finally, we improve and enhance our corpus in two different ways. First, we detect approximately-matching parallel passages in the entire corpus. This turns out to be a computationally challenging task, but yields a very large number of parallel passages. After excluding 18.6 million words of frequently recurring passages within the corpus, we proceeded to compare each of the texts against the entirety of the corpus. Our initial run compared one third of the corpus (over 308 million words) with the rest of the corpus and yielded more than 5 million pairwise matches of passages over 20 words in length. We shed some light on the nature of parallel passages with our analysis. Second, we develop a simple text dating algorithm based on language modeling in order to date the large portion of undated texts in the corpus (1,200 texts). We validate the text dating quality both quantitatively and qualitatively.  

The remainder of this paper is organized as follows. In the next section we review related work on Arabic corpora. We then describe the initial corpus preparation (Section~\ref{sec:corpus-basic}) and our enhancements (Section~\ref{sec:corpus-enhancement}). We demonstrate the application of this corpus with several case-studies in Section~\ref{sec:applications} before concluding with ideas for future work.  
\section{Related Work}
\label{sec:related-work}
Though there has been increasing interest in compiling Arabic corpora in the past decade, very little work has been done on compiling historical corpora reflecting the long history of the Arabic language.  Most of the existing corpora focus on modern written Arabic texts, particularly online print media, though there are a growing number of corpora which feature written and to a lesser degree spoken material from Arabic dialects. We mention here several relevant corpora and refer to other surveys for more details~\cite{Al-Sulaiti:2004,zaghouani2014critical,shoufan-alameri:2015:WANLP,Al-Thubaity:2015:ACK:2812480.2812508}.

To date, there is only a small number of diachronically oriented corpora of Arabic. The King Saud University Corpus of Classical Arabic (KSUCCA)~\cite{alrabiah2013design}\footnote{\url{http://ksucorpus.ksu.edu.sa}} 
consists of approximately 50.6 million words from the first 4 Islamic centuries. It has been morphologically analyzed with the MADA tool~\cite{Habash:2005,Habash:2009}. 
Almost all of the texts are derived from the Shamela corpus. Text metadata is by century, so more granular buckets are not possible in the current state of this corpus. Other Classical Arabic corpora that are worth mentioning include a 5 million word corpus by ~\newcite{elewa2004}, which doesn't seem to be publicly available, a 2.5 million word corpus by~\newcite{Rashwan:2011:SAD:2209821.2210698},\footnote{\url{http://www.RDI-eg.com/RDI/TrainingData}} and Tashkeela, a 76 million word corpus of texts  from  Al-Maktaba Al-Shamela website.\footnote{\url{https://sourceforge.net/projects/tashkeela}} All these corpora are rather small and lack temporal metadata.


Finally, a few large corpora are available only via online search interfaces:
KACST Arabic Corpus~\cite{Al-Thubaity:2015:ACK:2812480.2812508} has more than 700 million words, including around 16 million words from the beginning of the Islamic era.
The Leeds Arabic Internet Corpus\footnote{\url{http://corpus.leeds.ac.uk/internet.html}} and the International Corpus of Arabic\footnote{\url{http://www.bibalex.org/ica/en/About.aspx}} contain 300 and 100 million words, respectively, but they include mostly modern texts. The well-known ArabiCorpus\footnote{\url{http://arabicorpus.byu.edu}} has more than 170 million words from diverse periods of time, and arTenTen~\cite{Arts2014357} is a 5.8 billion word web corpus, with a sub-corpus of 115 million words available through Sketch Engine~\cite{Kilgarriff04thesketch}. \newcite{Milichka14} mention 
CLAUDia
, also based on Shamela, but with added genre metadata; however, only a subset appears to be accessible via a web interface.\footnote{\url{http://arabiccorpus.com/index.htm}} While these corpora are very large and may contain texts from different periods, they are not directly accessible and also lack sufficient diachronic information. 

In contrast to previous resources, our corpus has fairly fine-grained time information, it covers most of the history of the written Arabic language, and it is available for developing NLP applications or supporting digital humanities projects.

\section{Initial Corpus Preparation}
\label{sec:corpus-basic}

\subsection{Metadata/data wrangling}
Al-Maktaba Al-Shamela (``The Complete Library'') is a website which collects and stores digitized copies of important texts from throughout the history of the Arabic language. Sponsored by a religious charity, 
the texts are largely religious in nature, though there was not a strong division between religious and secular texts in the pre-modern era, and so many of the pre-modern texts are simply part of the Islamicate intellectual and literary tradition. The project itself is primarily designed to be a resource for reading individual texts --- it is not designed as a corpus per se. Most of the texts available on the website have been digitized, largely by manual double-keying, though some have been automatically digitized and are marked as such within the website's metadata. Texts are digitized from specific print editions, which, in most cases, allows for accurate citation in scholarly works. Texts can be accessed directly through the website, but there is also a Windows-based proprietary application which allows for off-line access. 

 The EPUB versions of the documents were downloaded from Shamela website and converted into specially designed markup format which allows for rapid manual editing as well as automated processing.\footnote{\url{https://github.com/maximromanov/mARkdown}} 
 One of the major challenges was inconsistent metadata, which was cleaned through automatic grouping of author names, book titles, and document distance using a custom script\footnote{\url{https://github.com/maximromanov/DuplAway}} to suggest matches which were then manually resolved. 
With most duplicates identified, records with more complete metadata were used to fill in the gaps in records where metadata was missing. We used Python scripts to further reconcile the metadata file with the Access databases included with the desktop application. 
 Numeric author codes from the database were integrated into the metadata file where possible, as it can be important to try to investigate a feature on an author-by-author basis rather than on a text-by-text basis. Word counts were obtained and integrated as well. 

\subsection{Lemmatization}
The raw corpus with accompanied metadata is already a useful resource for historical analysis. However, the orthography and morphology of the Arabic language pose several challenges for such research. First, the rich morphology leads to multiple surface forms for each common lemma, or dictionary entry. Second, orthographic norms in writing Arabic, such as the usual omission of diacritics, introduce a high degree of ambiguity. These phenomena hinder the ability to exploit Arabic corpora for lexical and semantic studies. To mitigate these problems, we automatically analyzed the entire corpus with MADAMIRA~\cite{PASHA14.593.L14-1479}, a state-of-the-art morphological analyzer and disambiguator. Given an Arabic sentence, it performs orthographic normalization (e.g. of many variant forms of the letter Alif), morphological analysis, and context-aware disambiguation. The result is a full analysis per word, including tokenization, lemmatization, part-of-speech-tagging, and various morphological features.

After lemmatization, we observe that the word vocabulary size in the corpus is about 16.8 million words, whereas the lemma vocabulary size is only 95 thousand lemmas. The lemmatized corpus will be made available to the research community, with future versions including other morphological features. 

\subsection{Corpus statistics and characteristics}

\begin{figure}[t]
\begin{subfigure}{0.4\textwidth}
\centering
\includegraphics[width=\textwidth]{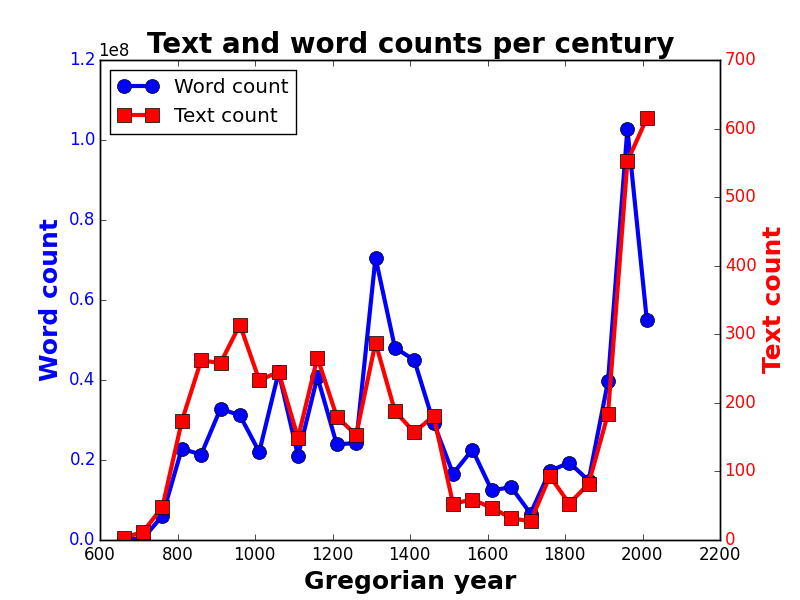}
\caption{Word and text counts per 50 year period.}
\label{fig:count-words-texts}
\end{subfigure}
\hspace{\fill}
\begin{subfigure}{0.4\textwidth}
\centering
\includegraphics[width=\textwidth]{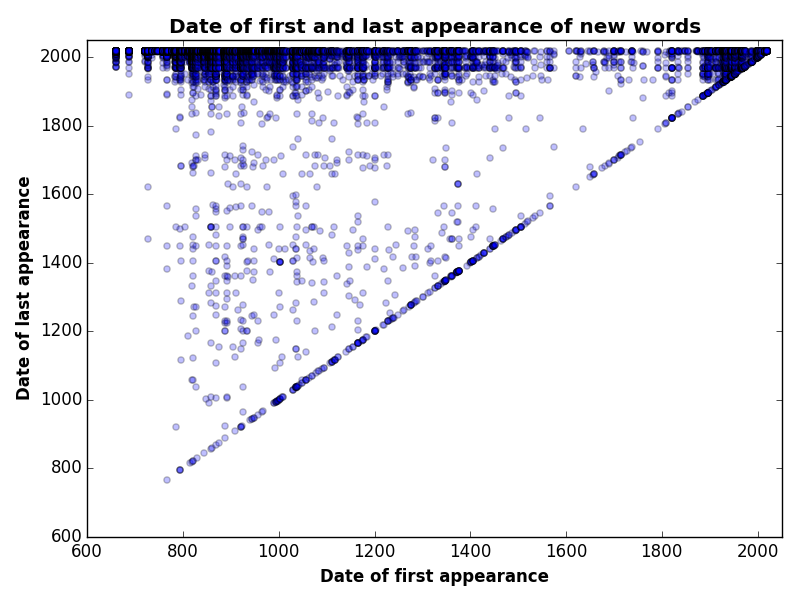}
\caption{The Arabic word life span, showing first and last usage date for every word lemma.}
\label{fig:lifespan}
\end{subfigure}
\caption{Word and text counts (left) and word lifespans (right) in the corpus.}
\label{Label}
\end{figure}

The dataset contains more than 6,100 texts, of which a significant portion (about 1,200) are undated. In the next section we consider automatic dating for these texts. Excluding undated texts, the corpus contains roughly 4,900 texts totaling 800 million words.

\begin{wraptable}{R}{0.4\textwidth}
\begin{minipage}{\textwidth}
\vspace{-2.3em}
\hspace{-1em} \small
\begin{tabular}{|l|l|l|}
\hline
Genre & Date (H/CE) & Texts \\
\hline
Hadith Collections & 335/946 & 179 \\
Biographies & 735/1334 & 377 \\
Jurisprudence (\textit{Fiqh}) & 891/1486 & 157 \\
Popular religious writing & 1419/1998 & 298 \\
\hline
\end{tabular}
\vspace{-1em}
\caption{Average date and total number of texts for example genres.}
\label{tab:data-genres}
\end{minipage}
\end{wraptable}

The texts are almost exclusively in formal Arabic, due to the religious focus, with a small number of contemporary texts showing colloquial elements. They are dated based on the author's date of death (DOD), with the earliest text dated to 41/661 Hijri/CE.\footnote{The texts are dated by the Islamic calendar (`H', for \emph{hijriyy}), which we convert to Gregorian dates (`CE').  We coded 
texts by living authors  with the date 1440 H/2018 CE to distinguish them from authors with a 1434-1435/2013 CE DOD. }
Figure~\ref{fig:count-words-texts} shows the distribution of word and text counts per century in the corpus, and Table~\ref{tab:data-genres} provides some statistics for the most common genres in the dataset. These genres are exemplary of the changes in dominant topics of 
discussion throughout the history of the Islamic world, with the collection of \emph{hadith}, sayings of the prophet Muhammad, a major concern in early centuries as the living links to hadith reciters were still present, while biographical works are developed more during an age of encyclopedism in the Middle Ages \cite{Muhanna14Encycl}. As literacy has become more widespread, popular religious writing has become a major genre. Thus there is an interaction of date and genre, though overall the genres of the corpus are relatively similar in their religious focus --- i.e. there are few of the modern secular texts that one finds in corpora based on media. 

The nature of writing in religious texts such as the ones in our corpus presents several challenges. 
First, authors tend to quote or paraphrase earlier texts, occasionally copying large chunks of texts. Second, many classical texts include contemporary introductions that are written in MSA. We touch upon the first issue in the next section, but leave a more systematic treatment of such problems for future work. 

\section{Corpus Enhancements}
\label{sec:corpus-enhancement}

\subsection{Text reuse and duplication}
A major desideratum in approaching this corpus is being able to detect, and potentially eliminate duplicate and reused text segments. Quotation of extensive sections of material is common in the Islamicate literary tradition, but unlike modern quotation it is not distinguished from the text at large. Identification of reused text is valuable for computational linguistic work, and for DH projects. 


Previous work on text reuse addressed the problem in the context of domains such as law bills or newspaper texts~\cite{Smith:2014:DML:2740769.2740800,wilkerson:2015}. Much of the previous work relies on n-grams to align similar chunks of texts; we refer to~\newcite{Smith:2014:DML:2740769.2740800}; \newcite{li:2016} for more details.\footnote{A notable system is \texttt{passim}, available at: \url{https://github.com/dasmiq/passim}. For a pilot application to Arabic, see \url{http://kitab-project.org/kitab/index.jsp}.} An especially interesting study is~\newcite{zemanek-milivcka:2014:CLFL}, which detected quotations in the CLAUDia corpus (Section~\ref{sec:related-work}) and built a network of documents based on metadata and quoted texts. However, their method focuses on long, verbatim quotations, whereas we are interested in approximately-matching parallel passages with possible variations. A standard approach to approximate-matching tasks is the use of edit-distance measures such as Levenshtein Distance; however, such an approach is unfeasible given a corpus of this size.
Instead, we follow a recent approach introduced by~\newcite{shmidman2016identification} for finding parallel passages in a Hebrew/Aramaic corpus. This method is appealing to use in our case for two reasons: the similarity between Arabic and Hebrew and the very efficient algorithm that can handle such a large corpus. We briefly review their approach and then describe our adaptation of the method. 

In~\newcite{shmidman2016identification}, every word in the corpus is represented by a two-letter hash, containing the two least common-letters from within the word. Then, for every position in the text, the subsequent 5 words are represented by four separate skip-grams, each one omitting a different word. These hashes and skip-grams allow efficient hash-based identification of matching passages while allowing for variants in orthography, differing prefixes and suffixes, and interpolated or omitted words. However, in order to apply this algorithm to the Shamela corpus, we first needed to remove the many "boiler plate" sentences and paragraphs which recur dozens, hundreds, or thousands of times within the corpus. These numerous repetitions would otherwise cause the sets of matching skip-grams to expand to unwieldy sizes.

We ran a preprocessing procedure to identify and mark all such phrases over the entire corpus (815 million words). It took two hours on a 32-CPU machine, marking 18,661,633 words (about 2 percent of the corpus) as part of frequently-recurring passages. After excluding those passages, we ran the full skip-gram algorithm.  As of this writing, the skip-gram algorithm completed over one third of the corpus, running for about a week on a 128-CPU machine and outputting more than 5 million pairwise matches of passages over 20 words in length, with an average length of 40 words. 

A manual inspection of the results shows that the text reuse  algorithm is extremely promising. The preprocessing step largely identifies formulaic prayers and Quranic verses. Occasionally, it captures lengthy sayings of the prophet (\emph{hadith}) when they are repeated frequently enough in the corpus. The results of the main run still included many sayings of the prophet, since they are an extremely important topic in Islamic thought, but almost no simply formulaic utterances. Furthermore, the repeated segments from the main run are more indicative of quotation. As one example among many, we are able to track the quotation of a single paragraph-length biography of a rather disreputable sheikh from its earliest quotation in 1359 CE, to another text in 1437 CE, to another in 1505 and finally to a modern text.

\subsection{Text dating}
While most of the texts in our corpus are dated by the author's date of death (Section~\ref{sec:corpus-basic}), a large portion has no associated date. In most cases, the date metadata was simply never entered, but can usually be verified by manual inspection of the metadata records (author information is often in prose form and includes dates) or the text itself. Here we consider how to automatically date undated texts in our corpus. 

There is a fairly large body of work on text dating, especially using clues like time expressions, but also various other features~\cite{dalli-wilks:2006:ARTE,chambers:2012:ACL2012,niculae-EtAl:2014:EACL2014-SP,popescu-strapparava:2015:SemEval}. Previous research operated at different granularity levels and algorithmic methods, including pairwise learning-to-rank and multi-class SVMs~\cite{niculae-EtAl:2014:EACL2014-SP,popescu-strapparava:2015:SemEval}. Here we choose a simple approach to text dating, based on language models, which were also used by~\newcite{jong2005temporal}, although in a different way.




We formulate our dating problem as a ranking task. Given an undated text, we would like to generate a ranked list of candidate dates. This formulation can facilitate subsequent manual inspection of the texts. It can also be useful for language technology applications that require an approximate date.


\begin{wrapfigure}{R}{0.4\textwidth}
\includegraphics[width=\textwidth]{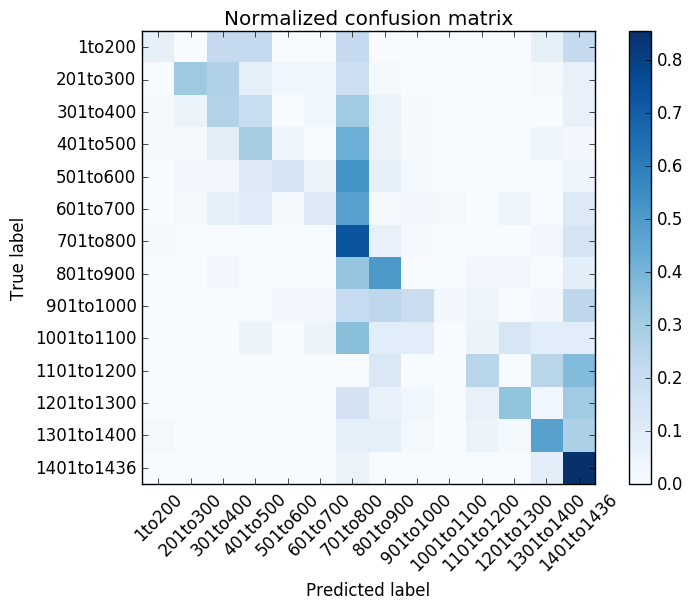}
\caption{Text dating confusion matrix.}
\label{fig:cm}
\end{wrapfigure}

We randomly split the dated texts in our corpus into train and test sets, containing 80\% and 20\% of the texts, respectively. This resulted in about 4500 texts for training and 900 for testing. We excluded dictionary documents from this experiment, as they tend to contain word items from various periods in time and can obscure the results. In practice, we bin the range of dates into buckets of 100 years to avoid sparsity.\footnote{Our Hijri-dated bins are: 1-200, 201-300, 301-400, ..., 1301-1400, 1401-1436, where we merge the 1-100 bin with 101-200 because it only contains 3 texts; the last bin runs from 1401 to the latest year in our corpus.} Then, for every 100 years range in our training corpus, we build a 5-gram language model with Knesser-Nay smoothing, using the SRILM toolkit~\cite{stolcke2002srilm,stolcke2011srilm}. Finally, for each text in our test corpus, we evaluate all trained language models, record their perplexity scores, and rank the predicted date ranges based on increasing perplexity values.

Table~\ref{tab:dating} shows the results of our text dating experiment, measured in accuracy at the top-$k$ predicted dates. The standard accuracy (at top-$1$) of the algorithm is 42.95, far above a random baseline of 7.14 and a majority baseline of 20.29. The correct result is found at the top-3 predictions over 70\% of the time, which we find encouraging for facilitating future dating of the undated portion of the corpus. 


\begin{table}[t]
\footnotesize
\setlength{\tabcolsep}{5pt}
\begin{tabular}{l|lllllllllllllllllllll}
$k$ & 1 & 2 & 3 & 4 & 5 & 6 & 7 & 8 & 9 & 10 & 11 & 12 & 13 & 14\\ 
Acc & 42.95 & 60.09 & 71.14 & 78.24 & 83.65 & 88.39 & 92.00 & 94.70 & 96.05 & 96.96  & 98.08 & 98.99 & 99.66 & 100.00 \\ 
\hline
\end{tabular}
\caption{Accuracy@$k$ of the text dating algorithm.}
\label{tab:dating}
\end{table}

Figure~\ref{fig:cm} shows the confusion matrix of the text dating algorithm. Most of the confusion occurs between subsequent date ranges, with two exceptions: the vertical bands at 701to800 and 1401to1436,  indicating that texts from other periods are often wrongly predicted as belonging to these two periods. It may be that texts from these periods use more diverse language, which is reasonable at least for the modern period (1401to1436 Hijri, turn of the 20th century) given the revival of the Arabic language in modern times. However, it is more likely an artifact of a larger number of texts  from these periods (Figure~\ref{fig:count-words-texts}), leading to better language models. Controlling for the size of the texts in each period is an interesting direction that we leave for future work.

Having confirmed the validity of this approach, we trained similar language models of 100-year bins from the entire dated corpus (training+testing). Then, we used these models to date all the undated texts. We manually examined around 10\% of the automatically dated texts, chosen randomly, and the results reflect those shown in Figure \ref{fig:cm}. In general, when a text is single-authored, the highest ranked candidate date is typically correct; if the two candidate dates are adjacent, it is almost definitely from within that period. However, many undated texts are compilations of one kind or another, and therefore it is difficult to assign an exact date of authorship that would align with the language used in the text. For example, one text having the 9th century as the first and the 15th century as the second candidate date is actually a reorganization of a 9th century text with a 15th century commentary. 
The interwoven nature of these texts suggests that the most productive annotation procedure will be to use the confusion index as a tool to prioritize manual tagging, a sort of tagging triage. Since the corpus has more than adequate coverage of modern texts, we are primarily interested in increasing pre-modern text coverage and so the automatic dating algorithm can be used to prioritize manual tagging of texts that are more likely to be pre-modern.

\section{Corpus applications}
\label{sec:applications}
In this section we give examples of how this corpus can be utilized for digital humanities applications.

\subsection{Digital Humanities}
The Shamela corpus, distributed privately, has been used in a number of projects in the digital Islamic humanities, though not all of these projects have reached publication stage.\footnote{Many of the presentations and workshops at Brown University's recurring Islamic Digital Humanities Workshops have made use of the Shamela corpus: \url{https://islamichumanities.org}} Material from Shamela and similar text collections have been combined with computational methods to explore which eras biographical chroniclers were interested in, the importance of geographical locations across the history of the Islamicate empires, and even to develop pedagogical materials using frequency to determine the ease of foreign language readings.\footnote{See \url{http://maximromanov.github.io} and \newcite{romanovdiss}} 

\subsection{Linguistics}
The Shamela corpus also represents an important resource for exploring the history of Standard written Arabic. Two brief case studies show the quantitative and qualitative analysis made possible by this corpus. Traditionally, questions about the history of Arabic are studied through impressionistic textual analysis, typically with no quantitative data. In this section we illustrate how our corpus can be used to provide a more objective, data-driven answer for these kinds of questions. 

\paragraph{The lifespan of Arabic words}
Standard written Arabic (SA) is the language of writing across the Arabic world and was standardized very early in the Islamic period. Qualitatively, there appears to be very little variation between modern written Arabic and pre-modern written Arabic from any era. Indeed, native speakers of Arabic do not regularly distinguish between SA in different eras, referring to both as ``eloquent Arabic." To a certain degree, a highly educated native speaker of Arabic should be able to approach texts from throughout the history of Arabic writing with significantly greater ease than an English speaker approaching pre-modern texts. We can use the Shamela corpus to check the intuition that there is a quantitative difference in the development of Arabic writing and the development of English. 



\begin{figure}[t]
\captionsetup[subfigure]{position=b}
\begin{subfigure}{0.45\textwidth}
\centering
\includegraphics[width=\textwidth]{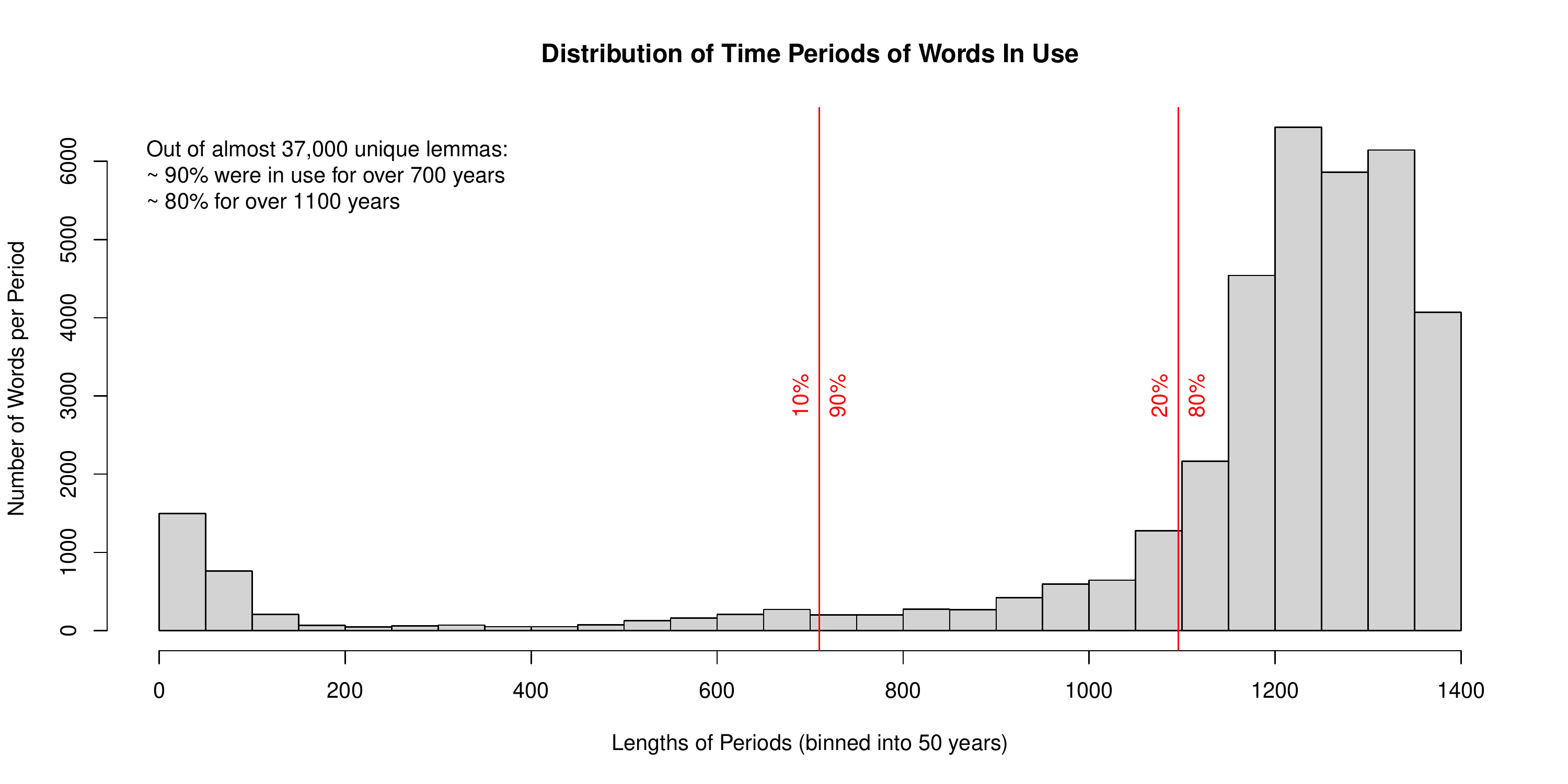}
\caption{Distribution of Arabic word lifespans.}
\label{fig:arabic-lifespan}
\end{subfigure}
\hspace{\fill}
\begin{subfigure}{0.45\textwidth}
\centering
\includegraphics[width=\textwidth]{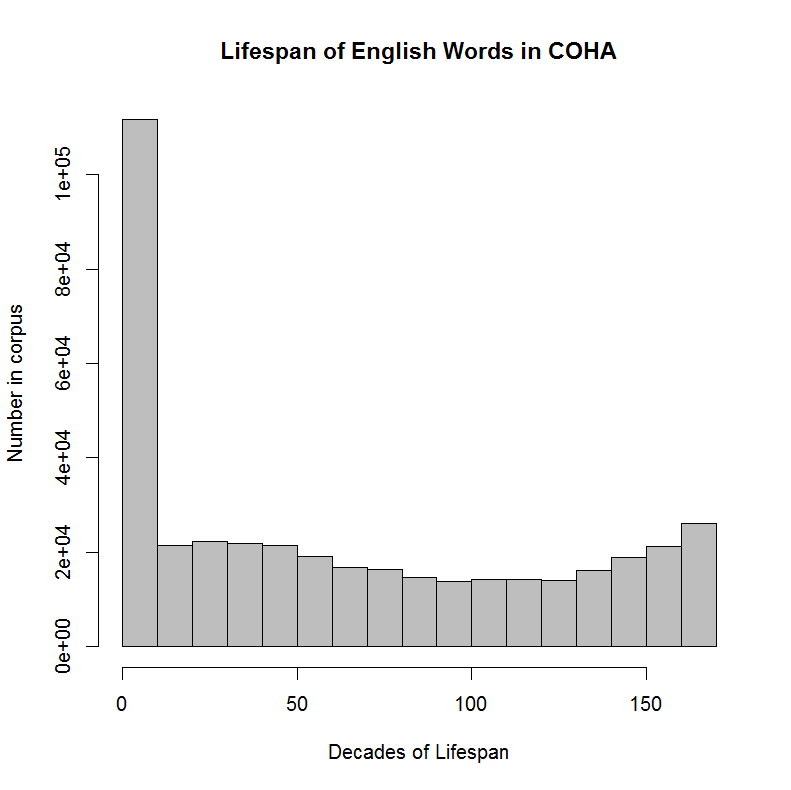}
\caption{Distribution of English word lifespans.}
\label{fig:english-lifespan}
\end{subfigure}
\caption{Distribution of word lifespans in Arabic (left) and English (right).}
\label{fig:ar-en-lifespans}
\end{figure}

To do this, we track the ``life'' of an Arabic word: for every word in the corpus, we find its first and last chronological usages.\footnote{In order to focus on word meaning, rather than morphological variation, we work with word lemmas from our lemmatized version of the corpus throughout this section.} As Figure~\ref{fig:lifespan} shows, Arabic words tend to have a very long life span --- words from all eras are still current. 
We can compare this to the Corpus of Historical American English (COHA)~\cite{davies2010corpus}. In the Shamela corpus the average Arabic word lifespan is 1124 years (SD: 338 years, median: 1222), about 83\% of the time span of the entire corpus. In  COHA, the average English word lifespan is 68 years (SD: 58, median: 60), about 36\% of the overall time span of the corpus. 
The difference between Arabic and English word lifespans is also demonstrated in Figure~\ref{fig:ar-en-lifespans}: Arabic words tend to stay in use for relatively much longer periods of time than English words.  

\paragraph{First Attestations}
A frequent task in historical lexicography is to investigate first attestations of words, or of particular uses of words. The techniques mentioned above have been used to find absolute first uses, but it is necessary to look at word contexts to investigate how words change. We consider a common linguistic hygiene genre that takes the form of `say/don't say' statements, usually implying that a modern usage of a word is inauthentic and not attested in earlier texts. Texts in this genre often circulate as memes, though there are also published works. Since it basically catalogs lexical innovations, this genre can be very helpful for finding words which have changed in meaning over time. To illustrate the application of qualitative research within the corpus, we report our findings for one claim of change found in a widespread online `meme'  which features several supposed common errors.\footnote{The meme is extremely widespread. An example instance can be found here: \url{http://www.inciraq.com/pages/view_paper.php?id=200914073} (accessed 28.4.2016)} The meme argues that the word \hawalay\
`around, approximately' is incorrectly used for approximation of number, though the source offers no guidance on how that word should be used.

To determine what the `original' meaning of \hawalay\ was and when it changed into its innovative meaning, we  ran a concordancing algorithm across the entire corpus. The concordance was sorted by date, and visually inspected to determine the original usage of the word, and to determine when and how it changes. We found that in early attestations \hawalay\ is used for physical approximation, referring to objects which physically surround or which are placed around or near a central location. Only later does it develop a numerical approximation function. However, the change in meaning happened quite early, and is not a modern innovation over Classical Arabic, as implied by the `say/don't say' meme.   The first instance of use as a numerical approximation occurs in a text whose author died in 1201 CE, i.e.  well before the modern era \cite[vol. 17: 285]{JawziAlMuntazam2}. The Shamela corpus thus allows us to rapidly investigate the histories of individual words using the traditional tools of linguistic investigation. 


\section{Conclusion and Future Work}
\label{sec:conclusion}

In this work we described our efforts to develop a large-scale, historical Arabic corpus, comprising 1 billion words from a 14-century time span. We also improved the quality of this corpus by automatic text dating and reuse detection, and demonstrated its utility for digital humanities and historical research. 

In future work, we aim to delve deeper into the history of the Arabic language and its possible periodization. We would also like to investigate mutual sources of influence between different scholars by analyzing automatically extracted parallel passages. Finally, we hope this corpus will serve the NLP/DH communities in promoting better understanding of the Arabic language and the culture of its speakers.


\bibliographystyle{acl}
\bibliography{coling2016}

\end{document}